\documentclass[default]{sn-jnl}

\usepackage{lineno,hyperref}
\usepackage{times}
\usepackage{url}
\usepackage{latexsym}
\usepackage{xcolor} 
\usepackage{soul}
\usepackage{microtype}
\usepackage{graphicx}
\usepackage{booktabs} 
\usepackage{microtype}
\usepackage{latexsym}
\usepackage{float}
\usepackage{multirow}
\usepackage{enumitem}
\usepackage{tabularx}
\usepackage{arydshln}
\usepackage{booktabs,fixltx2e}
\usepackage{amssymb,mathtools}
\usepackage{eqnarray,amsmath} 
\usepackage{hyperref}
\usepackage{microtype}
\usepackage{subfig}
\usepackage{pbox}
\usepackage{pifont}
\modulolinenumbers[5]
\usepackage[normalem]{ulem}
\usepackage{algorithm}
\usepackage{algpseudocode}
\definecolor{LightGoldenrod11}{rgb}{0.98, 0.98, 0.82}

\jyear{2023}%

\theoremstyle{thmstyleone}%
\theoremstyle{thmstyletwo}%

\theoremstyle{thmstylethree}%

\begin{document}
\nocite{*}
\title[Interpretable Multimodal Emotion Recognition using Hybrid Fusion]{Interpretable Multimodal Emotion Recognition using Hybrid Fusion of Speech and Image Data}


\author*[1]{\fnm{Puneet} \sur{Kumar}}\email{pkumar99@cs.iitr.ac.in}
\equalcont{These authors contributed equally to this work.}
\author[2]{\fnm{Sarthak} \sur{Malik}}\email{sarthak\_m@mt.iitr.ac.in}
\equalcont{These authors contributed equally to this work.}
\author[1]{\fnm{Balasubramanian} \sur{Raman}}\email{bala@cs.iitr.ac.in}

\affil[1]{\orgdiv{Department of Computer Science and Engineering}, \orgname{Indian Institute of Technology Roorkee}}

\affil[2]{\orgdiv{Department of Electrical Engineering}, \orgname{Indian Institute of Technology Roorkee}}


\abstract{This paper proposes a multimodal emotion recognition system based on hybrid fusion that classifies the emotions depicted by speech utterances and corresponding images into discrete classes. A new interpretability technique has been developed to identify the important speech \& image features leading to the prediction of particular emotion classes. The proposed system's architecture has been determined through intensive ablation studies. It fuses the speech \& image features and then combines speech, image, and intermediate fusion outputs. The proposed interpretability technique incorporates the divide \& conquer approach to compute shapely values denoting each speech \& image feature's importance. We have also constructed a large-scale dataset (IIT-R SIER dataset) consisting of speech utterances, corresponding images, and class labels, i.e., `anger,' `happy,' `hate,' and `sad.' The proposed system has achieved 83.29\% accuracy for emotion recognition. The enhanced performance of the proposed system advocates the importance of utilizing complementary information from multiple modalities for emotion recognition.}\keywords{}




\maketitle
\section{Introduction}\label{sec:intro}
The multimedia data has overgrown in the last few years, leading multimodal emotion analysis to emerging as an important research trend~\cite{baltruvsaitis2018multimodal}. The need to develop multimodal emotion processing systems capable of recognizing various emotions from images and texts is rapidly increasing. Research in this direction aims to help machines become empathetic as emotion analysis is used in various applications such as cognitive psychology, automated identification, intelligent devices, and human-machine interface~\cite{poria2017review}. The speech and image modalities portray human emotions and intentions very effectively~\cite{kim2018building}. Combining complementary information from both of these modalities could increase emotion recognition accuracy~\cite{zeng2009survey}.\vspace{.1in}

Researchers have attempted to identify emotions by processing audio and visual information separately ~\cite{xu2020improve,majumder2019dialoguernn,rao2019learning}. However, multimodal emotion recognition, where the emotional context from multiple modalities are analyzed together, performs better than unimodal emotion recognition~\cite{zeng2009survey}. In this context, multimodal emotion recognition from speech \& text modalities and image \& text modalities have been performed; however, emotion recognition from speech \& image modalities has yet to be fully explored. Moreover, most of the existing multimodal approaches do not focus on interpreting the internal working of their emotion recognition systems. It inspired us to develop a multimodal emotion recognition system capable of recognizing emotions portrayed by speech utterances \& corresponding images and explaining the importance of each speech segment \& visual feature towards emotion recognition.\vspace{.1in}

Multimodal emotion recognition also faces the issue of the unavailability of sufficient labeled datasets for training. Moreover, the real-life multimodal data contains generic images with facial, human, and non-human objects, but most of the existing multimodal datasets contain only facial and human images~\cite{busso2008iemocap}. A few multimodal datasets are available that contain generic images; however, they consist of positive, negative, and neutral sentiment labels and do not contain multi-class emotion labels~\cite{gaspar2019multimodal, Vadicamo_2017_ICCVW}. A new dataset, `IIT Roorkee Speech \& Image Emotion Recognition (IIT-R SIER) dataset,' has been constructed to address this issue. It contains generic images, corresponding speech utterances, and discrete class labels, i.e., `anger,' `happy,' `hate,' and `sad.' We used the data instances with identical predicted emotion labels for image and text modalities to construct the dataset. This paper analyses the improvements in SER on combining the complementary information from corresponding images.\vspace{.1in}

The proposed system, `ParallelNet,' recognizes emotions in speech utterances and corresponding images. It implements two networks, $N1$ and $N2$, to fuse the information of speech and image modalities in a hybrid manner of intermediate and late fusion. The architectures for $N1$ and $N2$ are determined through extensive ablation studies. A technique to interpret the important input features and predictions has also been developed. The proposed system has performed with an accuracy of 83.29\% on the IIT-R SIER dataset. The dataset and code for this paper are accessible at \href{https://github.com/MIntelligence-Group/SpeechImg_EmoRec}{https://github.com/MIntelligence-Group/SpeechImg_EmoRec}.\\~\\

\noindent The paper makes the following major contributions.\vspace{.05in}
\begin{itemize}
	\item A hybrid-fusion-based novel system, `ParallelNet,' has been proposed to classify an input containing speech utterance \& corresponding image into discrete emotion classes. It combines the information from speech \& image modalities using a hybrid of intermediate and late fusion.\vspace{.05in}
	\item A large-scale dataset, `IIT-R SIER dataset' containing speech utterances, corresponding images, and emotion labels, has been constructed.\vspace{.05in}
	\item A new interpretability technique has been developed to identify the important parts of the input speech and image that contribute the most to recognizing emotions.\vspace{.1in}
\end{itemize}

Further in this paper, the related works have been reviewed in Section~\ref{sec:lr}. The proposed dataset, system, and interpretability technique have been described in Section~\ref{sec:method} along with the dataset compilation procedure. Section~\ref{sec:experiments} and \ref{sec:results} discuss the experiments and results. Finally, Section~\ref{sec:conclusion} concludes the paper and highlights the directions for future research.

\section{Related works}\label{sec:lr}
\noindent This Section surveys the existing literature on speech \& image emotion recognition and the interpretability of deep neural networks.

\subsection{Speech emotion recognition}
The deep learning-based approaches using spectrogram features and attention mechanisms have shown state-of-the-art results for speech emotion recognition (SER) \cite{dai2019learning,yenigalla2018speech,kumar2021end}. In this context, Xu et al.~\cite{xu2020improve} generated multiple attention maps, fused and used them for SER. They observed an increased performance as compared to non-fusion-based approaches. In another work, Majumder et al. \cite{majumder2019dialoguernn} implemented a deep neural network to track speakers' identities showing specific emotions.

\subsection{Image emotion recognition}
Image Emotion Recognition (IER) research is also an active domain. For instance, Kim et al.~\cite{kim2018building} built a deep feed-forward neural network to combine different levels of emotion features obtained by using the semantic information of the image. In another work, Rao et al.~\cite{rao2019learning} prepared hierarchical notations for emotion recognition in the visual domain.\vspace{.1in}

The human emotions can be expressed in various modalities, out of which speech \& image express the emotional intentions most effectively~\cite{kim2018building}. Analysis in a single modality may not be able to recognize the emotional context completely, which leads to the need for multimodal emotion recognition approaches that analyze multimodal audio-visual emotional context~\cite{zeng2009survey}.

\subsection{Multimodal emotion recognition}
Multimodal emotion analysis from audio-visual data has started getting researchers' attention lately~\cite{hossain2019emotion,kumar2021hybrid,guanghui2021multi}. For instance, Siriwardhana et al.~\cite{siriwardhana2020jointly} fine-tuned Transformers-based models to improve the performance of multimodal speech emotion recognition. Multimodal emotion recognition has been carried out for text \& speech modalities
\cite{makiuchi2021multimodal,kumartowards} and text \& image modalities \cite{Vadicamo_2017_ICCVW,gaspar2019multimodal,kumar2021hybrid}. However, it has not been fully explored for speech \& image modalities. Moreover, most deep learning-based multimodal emotion recognition systems work as a black box where it is difficult to interpret their inside mechanism. It inspired us to develop an interpretable multimodal emotion recognition system for speech \& image modalities.

\subsection{Interpretability of deep neural networks}
The existing interpretability approaches compute each input feature's importance by backpropagating the network or observing the changes in output on changing the input~\cite{lundberg2017unified}. In this direction, Riberio et al.~\cite{ribeiro2016should} explained a network based on each input's importance. Researchers have explained the layer-by-layer learning of deep neural networks and the output based on all the neurons' contributions~\cite{kumartowards,shrikumar2017learning}. There are interpretability methods for visual analysis to compute input pixels' importance~\cite{lundberg2017unified,ribeiro2016should,malik2021towards}. However, such methods still need to be sufficiently explored for speech modality. It inspired us to develop an interpretability technique for multimodal emotion recognition to explain the importance of each speech segment and each visual feature of the input.

\section{Proposed Methodology}\label{sec:method}
\subsection{Dataset construction}\label{sec:method_data}

The `IIT Roorkee Speech \& Image Emotion Recognition (IIT-R SIER) dataset has been constructed using Balanced Twitter for Sentiment Analysis (B-T4SA) dataset~\cite{Vadicamo_2017_ICCVW}. The recent text-to-speech models generate high-quality audio that can be used as a valid approximation of natural audio signals \cite{ping2018deep,deepmind2016wavenet,oord2016wavenet}. A pre-trained state-of-the-art text-to-speech model, DeepSpeech3~\cite{ping2018deep}, has been used to convert the text from the B-T4SA dataset to speech. The samples are manually cleaned by removing the corrupt and duplicate samples. Further, the following procedure has been followed to generate the ground-truth labels according to the overall emotional context represented by both modalities in combination. \vspace{.1in}

\begin{figure}[!h]
	\centering
	\includegraphics[width=0.54\textwidth]{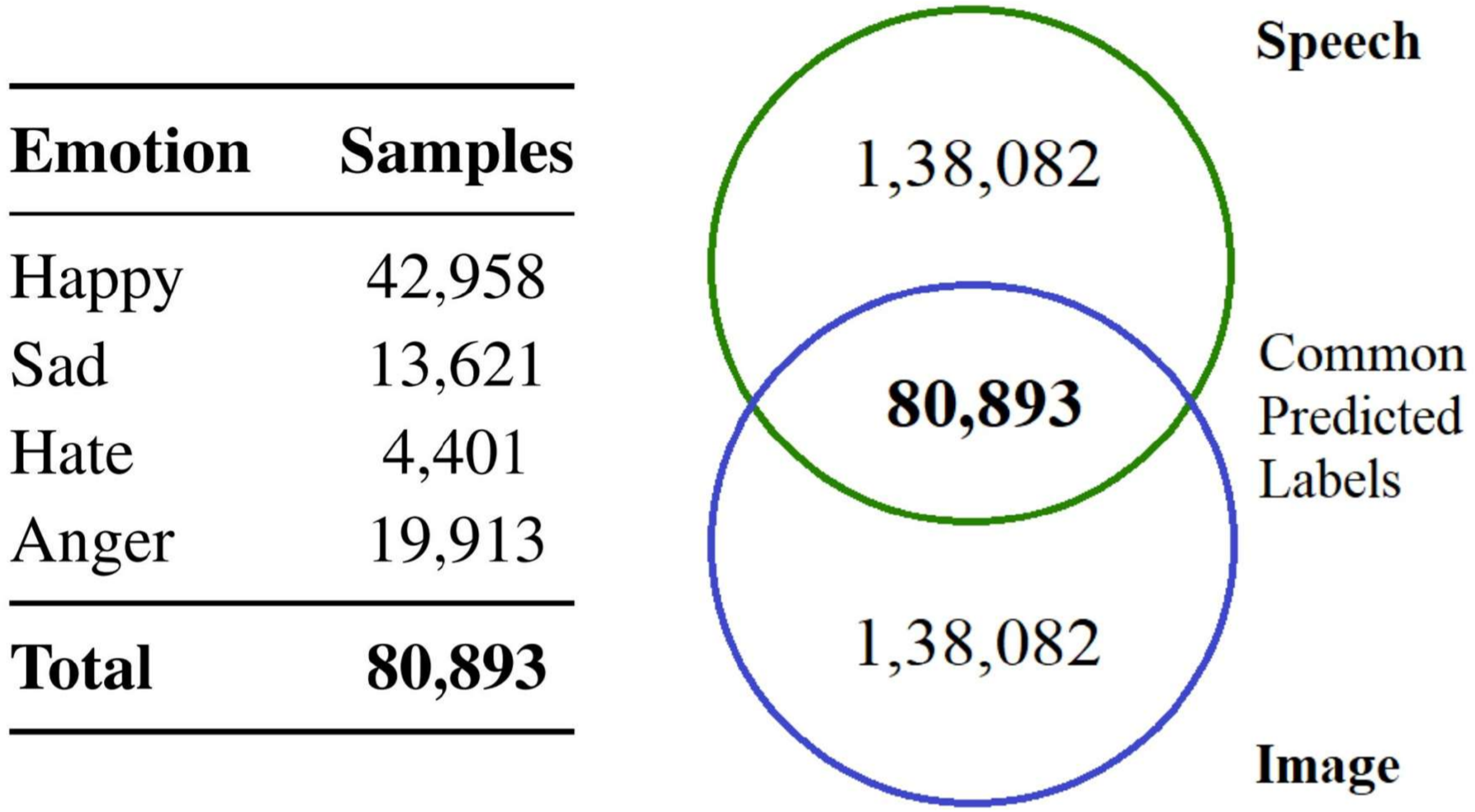}\vspace{.07in}
	\caption{Summary of IIT-R SIER dataset. Left: Class-wise data samples distribution. Right: Modality-wise data distribution.} 
	\label{fig:data}
\end{figure}

Various parameters of the SIER dataset have been summarised in Fig.~\ref{fig:data} whereas the procedure to construct the same has been described as follows.\vspace{.1in}

The speech component of each data sample is passed through the SER model trained on The Interactive Emotional Dyadic Motion Capture (IEMOCAP)~\cite{busso2008iemocap} dataset; classification probabilities for each emotion class are obtained, and the maximum among the probabilities for all emotion classes is noted as $max_1$. Likewise, each sample's image component is passed through the IER model trained on Flickr \& Instagram (FI)~\cite{you2016building} dataset, and the maximum of classification probabilities for all emotion classes, i.e., $max_2$ is noted. The higher of $max_1$ and $max_2$ is observed, and the corresponding emotion label is assigned as the ground-truth label to the data sample. For example, if IER model returned probabilities $0.1$, $0.8$, $0.05$ and $0.05$ for four emotion classes while SER model gave $0.1$, $0.1$, $0.7$ \& $0.1$ then we assigned second emotion class to the sample considering $max$($0.8$ and $0.7$). The samples having $max$($max_1$, $max_2$) less than a threshold of $0.5$ are discarded as the predicted class label must be at least double confident than random prediction (probability 0.25). The samples labeled as `excitement' \& `disgust' have been re-labeled as `happy' \& `hate' as per Plutchik's wheel of emotions~\cite{plutchik2001nature}. The final dataset contains a total of 80,893 samples with 42,958 labeled as `happy,' 13621 as `sad', and 4401 \& 19,913 as`hate' and `anger' respectively.\vspace{.1in} 



We did not take the samples with the same predicted labels by SER and IER, as speech \& image modalities might favor different emotion classes in isolation. In contrast, we are interested in the emotion class denoted by both modalities together. Samples for which SER and IER models predicted the same emotion label have been retained to form the IIT-R SIER dataset. The samples having the same predicted labels for SER and IER models denote high confidence in both modalities. They have been retained irrespective of whether the labels are correct. This approach is inspired by the B-T4SA dataset's base paper, where the samples having high confidence in text and image modalities are kept while others are discarded~\cite{Vadicamo_2017_ICCVW}.\vspace{.1in} 

\subsubsection{Human evaluation}\label{sec:human_eval} 
We had two human readers (one male and one female) who spoke out and recorded the text components of the data samples. The evaluators listened to the machine-synthesized and human speech recorded by the human readers and labeled the emotion classes portrayed by them. The samples have been picked randomly, and the average of the evaluators' scores has been reported in Table \ref{tab:human_eval}. Here, $A_{i}$ denotes the emotion classification accuracy when the human evaluators predicted the emotions considering the image components. Likewise, $A_{ss}$ \& $A_{hs}$ are the accuracy values on considering the synthetic and human speech components, and $A_{ss-i}$ \& $A_{hs-i}$ are the accuracies on considering both speech and image modalities.

\begin{table}[!h]
\centering	
\caption{Human evaluation of SIER dataset. Where $A_{\textit{m}}$ denotes the emotion classification accuracy for modality $m$, $i$: image modality, $ss$: synthetic speech, $hs$: human speech, $ss-i$: multimodal context combining synthetic speech and image modalities and $hs-i$: multimodal context combining human speech and image modalities.}
\label{tab:human_eval}
\resizebox{.82\textwidth}{!}
{%
    \begin{tabular}{lccccc}\toprule
    \textbf{Class}   & \textbf{$A_{i}$} & \textbf{$A_{ss}$} & \textbf{$A_{hs}$} & \textbf{$A_{ss-i}$} & \textbf{$A_{hs-i}$} \\ \midrule
    \textbf{Happy}   \hspace{.15in} &63.89\% &66.67\% & 69.44\% & 73.48\% & 75.87\% \\
    \textbf{Sad}     \hspace{.15in} &75.00\% &77.08\% & 78.13\% & 82.43\% & 83.27\% \\ \textbf{Hate}    \hspace{.15in} &67.86\% &71.43\% & 72.32\% & 77.64\% & 81.32\% \\ 
    \textbf{Anger}   \hspace{.15in} &70.31\% &82.81\% & 85.94\% & 82.17\% & 84.19\% \\
    \hdashline
    \textbf{Overall} \hspace{.15in} &69.26\% &74.49\% & 76.46\% & 78.93\% & 80.46\% \\ \bottomrule
    \end{tabular}
}
\end{table}

The following two major observations can be drawn from Table \ref{tab:human_eval}: \textbf{i)} The similar values of 74.49\% for synthetic speech and 78.91\% for human speech advocate that the speech component of the data generated through text-to-speech is mature enough and embodies the appropriate emotional context. \textbf{ii)} Considering complementary information from both speech and image modalities led to higher emotion recognition performance. The evaluators also reported that 78.93\% of the samples considering machine-synthesized speech along with the corresponding image were in line with the determined emotion label, whereas this is comparable to the value of 80.46\% on considering human speech along with the corresponding image with is significantly higher than the accuracies on considering only image or only speech components. 

\subsection{Proposed multimodal emotion recognition system} 
\begin{figure}[!t]
	\centering
	\includegraphics[width=1\textwidth]{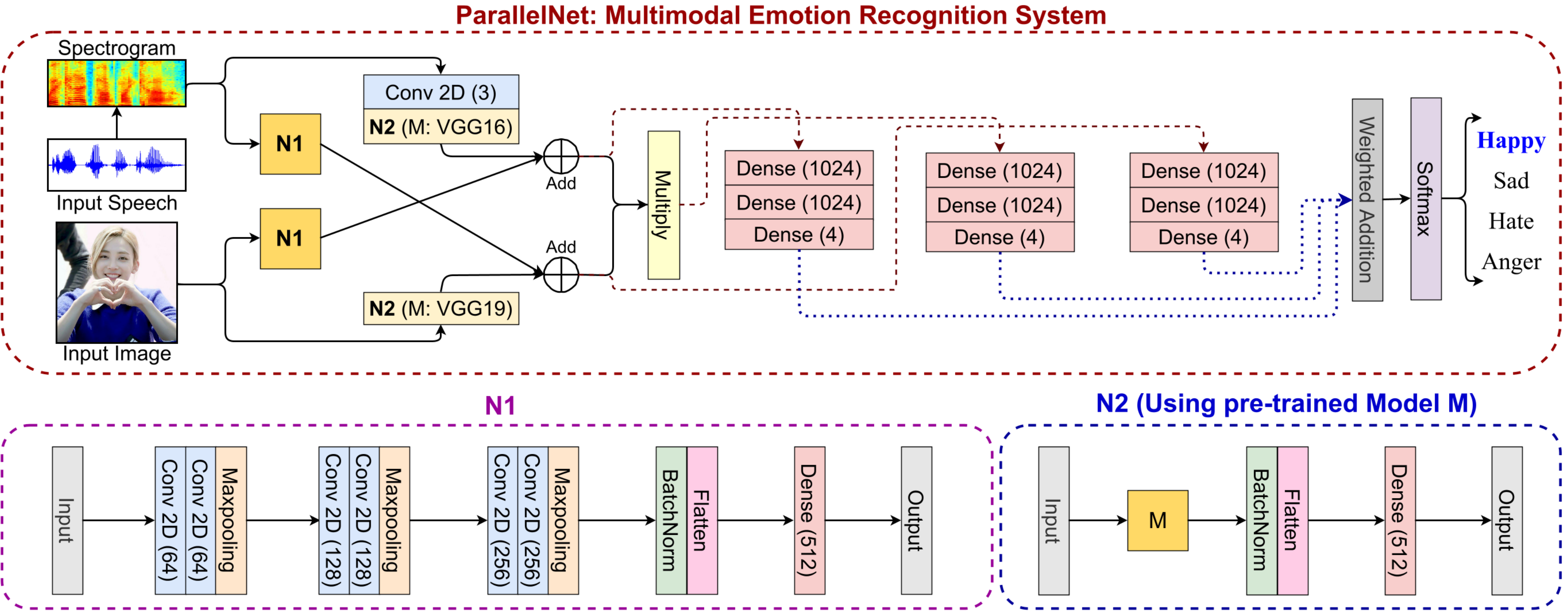}\vspace{.1in}
	\caption{Architecture of the proposed system (top), $N1$ (bottom left) and $N2$ (bottom right) where $M$ is a pre-trained model.} 
	\label{fig:archi}
\end{figure}

Fig.~\ref{fig:archi} depicts the architecture of the proposed multimodal emotion recognition system, which is determined in Section~\ref{sec:ablation} through the ablation studies. A hybrid of intermediate and late fusion is implemented where intermediate fusion combines various modalities' information before classifying, while late fusion fuses the results after classification. The input image is in the space domain. The speech has been converted from the time domain to a log-mel spectrogram, i.e., the space domain. The proposed system contains networks $N1$ and $N2$ and dense, multiply, weighted addition, and softmax layers. $N1$ uses convolution \& max-pool layers while $N2$ uses pre-trained networks VGG16 and VGG19~\cite{simonyan2014very}. Both of these networks contain batch-normalization, flattened, and dense layers. \vspace{.1in}

The intuition behind our architecture was to include a mechanism somehow So that each modality affects the other while making predictions. Here the two modalities are combined in two ways:- intermediate fusion and late fusion. First of all, to bring both modalities in the same domain audio signal is converted to a log-mel spectrogram to convert it from the time to space domain. Now, let us consider two networks, N1 and N2. N1 consists of a pre-trained network, than a batch normalization layer, a flattening layer, and a dense layer of 512 neurons. While N2 has the following architecture: First, two convolution layers have 64 filters, then a max-pooling layer, then two more convolution layers of 128 filters, then a max-pooling layer again, comes two more convolution layers of 256 filters, and then a max-pooling layer. Then it consists of a batch normalization layer, a flattening layer, and a dense layer of 512 neurons.

\subsubsection{Intermediate fusion phase}\label{sec:if}
Consider the networks fed with the image input to be $N1_i$ and $N2_i$ while the networks $N1_s$ and $N2_s$ process the speech input. The speech is expressed as a spectrogram of size (128, 128, 1) and passed first to a convolution layer having three convolutional filters of size (1, 1) each and then to $N2_s$ where a pre-trained VGG16 network is used. The image with size (128, 128, 3) is passed to $N2_i$ that uses a pre-trained VGG19 network. As shown in Eq.~\ref{eq:eq1a}, the output of $N1_s$ is added with the output of $N2_i$ to get $F_s$. Likewise, the outputs of $N1_i$ and $N2_s$ are added to obtain $F_i$. Then $F_s$ and $F_i$ are element-wise multiplied to obtain $F_{mul}$.\vspace{.1in}


{
	\begin{eqnarray}
	\label{eq:eq1a}
	\begin{split}	
	&F_i = Add(output(N1_i),\ output(N2_s))\\
	&F_s = Add(output(N1_s),\ output(N1_i))\\
	&F_{mul} = Multiply(F_i,\ F_s)\\
	\end{split}
	\end{eqnarray}
} 

The choice of using multiplication instead of weighted addition in Eq. \ref{eq:eq1a} to combine $F_s$ and $F_i$ in the low-level fusion has been determined experimentally. Moreover, theoretically, if the speech and image modalities predict the same emotion class, they should support each other. However, let's consider a case where one modality predicts $i^{th}$ emotion very strongly while another predicts another emotion $j^{th}$ weakly. We expect that the $i^{th}$ emotion should be predicted weakly. It would not have been the case in the case of using addition, and the $i^{th}$ emotion would have the upper hand. In comparison, the multiplication of both modalities would dilute the assertive behavior of the $i^{th}$ emotion and give us the expected prediction. 

\subsubsection{Late fusion phase}
The intermediate outputs $F_i,\ F_s,\, and\ F_{mul}$ are passed from three dense layers of size 1024, 1024, and 4 to obtain $O_{sp}$ for speech, $O_{img}$ for image, and $O_{mul}$ for multiplied. These outputs are combined using the weighted addition layer as per Eq.~\ref{eq:eq4} in a late fusion manner and passed from a softmax layer to get the final predicted label, $\hat{y}$. The weights $w_1,\ w_2,\ \text{and}\ w_3$ are randomly initialized and passed to a softmax layer to normalize them to non-negative values. Their final values are learned using the Gradient Descent algorithm. It combines the information from speech \& image modalities and the output of intermediate fusion in a hybrid manner.\vspace{-.1in}

{
	\begin{eqnarray}
	\label{eq:eq4}
	\begin{split}
	&O =  w_1 \times O_{sp} + w_2 \times O_{img} + w_3 \times O_{mul} \\
	&\hat{y} = Softmax (O)
	\end{split}
	\end{eqnarray}
}

\subsection{Proposed interpretability technique} 
While making predictions, a deep learning-based classifier is expected to consider the input features that a human would consider. However, it is challenging to look into it and understand what input features it is considering~\cite{ribeiro2016should}. To work on this challenge, we have developed an interpretability technique based on `shapely values'~\cite{lundberg2017unified} that denotes each input feature's importance. Theoretically, shapely values' computation takes exponential time. The computation has been approximated using the divide and conquer approach as shown in Eq. \ref{eq:eq2}. For a model with two features $f_1$ and $f_2$, shapely value $\mathscr{S}_{\{f_1\}}$ for feature $f_1$ denoting its importance is computed as follows.\vspace{-.1in} 

\begin{figure*}[!t]
	\centering
	\includegraphics[width=1\textwidth]{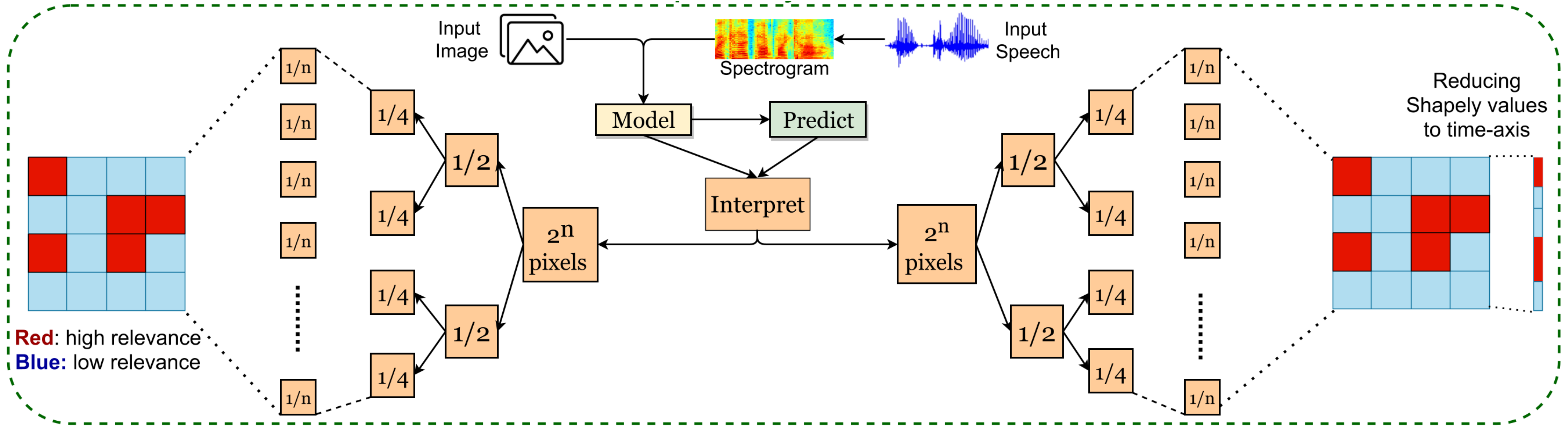}\vspace{.1in}
	\caption{Proposed interpretability technique's illustration. Here, each part's importance is computed using Divide \& Conquer.}
	\label{fig:archi2}
\end{figure*}

{
	\begin{eqnarray}
	\label{eq:eq2}
	\begin{split}	
	&\mathscr{S}_{\{f_1\}} = (1/2) \times MC_{f_1,\{f_1\}} + (1/2) \times MC_{f_1,\{f_1,f_2\}} \\
	\end{split}
	\end{eqnarray}
}

\noindent
Here, $MC_{f_1,\{f_1\}}$ is feature $f_1$'s marginal contribution to the model containing only $f_1$ and given by Eq. \ref{eq:eq3} where $score_{\{f_1\}}$ denotes the prediction for the ground-truth label using the model with feature $f_1$.

{\fontsize{10}{11}\selectfont 
	\begin{eqnarray}
	\label{eq:eq3}
	\begin{split}	
	&MC_{f_1,\{f_1\}} = score_{\{f_1\}} - score_{\{\phi\}}\\
	\end{split}
	\end{eqnarray}  
}

The respective speech and image inputs are segregated and fed into the model while keeping the other as zero to compute the individual contribution of each modality. As depicted in Fig.~\ref{fig:archi2}, each modality's input is divided into two parts for a specified number of times, and the importance of each part towards the model's prediction is computed as per Eq.~\ref{eq:eq4}. Moreover, the calculation of the importance score follows the basic requirement of shapely values given by Eq.~\ref{eq:eq5}.\vspace{-.1in}

{\fontsize{10}{11}\selectfont 
	\begin{eqnarray}
	\label{eq:eq5}
	\begin{split}
	& \mathscr{S}_{\{f_1\}} + \mathscr{S}_{\{f_2\}} = \mathscr{S}_{\{f_1,\ f_2\}} - \mathscr{S}_{\{null\}}\\
	\end{split}
	\end{eqnarray}
}

The important image features for the predictions can be directly observed through the shapely values. In contrast, the important speech features are analyzed after transforming them to wave, i.e., time-domain representation. We first applied the shapely values directly and converted the spectrogram to speech; however, the speech reconstructed by this method was not meaningful. Then, we used the method of averaging the shapely values along the frequency axis and reducing them to the time axis to find the features' importance at a given time. The speech segments below a threshold shapely values of 30 percentile have been reduced to zero. The leftover segment with high importance is converted to text using speech-to-text model~\cite{chan2016listen} and interpreted to understand how the model classifies each instance. The proposed interpretability technique has been summarised in Algorithm \ref{algo:a1}. 

{
\begin{algorithm}[]
	Define $model:$ Multimodal deep neural network\;\\
    Define $data\_img:$ Image pixels\;\\
    Define $data\_speech:$ Speech spectrogram pixels\;\\
    Define $wd,\ ht:$ Width \& height\;\\
    Define $times:$ Number of division to divide image \& speech spectogram in\;\\
    Define $SHAP\_value\_img:$ Image's shaply value\;\\
    Define $SHAP\_value\_speech:$ Speech's shaply value\;\vspace{.1in}
    
    Procedure DnCShap\_MM($model$, $data\_img$, $data\_speech$, $wd$, $ht$, $times$)~\vspace{.1in}
    
    $\triangleright$ {\color{gray} Initialize data with all zero entries, here; np: numpy}\\
    $data\_1$ = np.zeros([$wd,\ ht$, 3])\;\\
    $data\_2$ = np.zeros([$wd,\ ht$, 1])\;\\
    $data\_1$ = data\_1.reshape(1, $wd,\ ht$, 3)\;\\
    $data\_2$ = data\_2.reshape(1, $wd,\ ht$, 1);~\vspace{.1in}
    
    $\triangleright$ {\color{gray} Make original data ready to be fed into $model$}\\
    $data\_f\_img$ = $data\_img$.reshape(1, $wd,\ ht$, 3)\;\\
    $data\_f\_speech$ = $data\_speech$.reshape(1, $wd,\ ht$, 1);~\vspace{.1in}
    
    $\triangleright$ {\color{gray} Find the predicted label}\\
    $pred$ = $model$.predict ($data\_f\_img$, $data\_f\_speech$)\;\\
    $arg\_max$ = np.argmax($pred$); ~\vspace{.1in}
    
    $\triangleright$ {\color{gray} Find predicted probability with original data}\\
    $pred\_f$ = $pred$[0][$arg\_max$]~\vspace{.1in};
    
    $\triangleright$ {\color{gray} Find predicted probability with blank data}\\
    $pred\_b$ = model.predict($data\_1$, $data\_2$)[0][$arg\_max$];~\vspace{.1in}
    
    $\triangleright$ {\color{gray} Find predicted probability with only image modality}\\
    $pred\_1$ = model.predict($data\_f\_img$, $data\_2$)[0] [$arg\_max$]\;~\vspace{.1in}

    $\triangleright$ {\color{gray} Find predicted probability with only speech modality}\\
    $pred\_2$ = model.predict($data\_1$, $data\_f\_speech$)[0] [$arg\_max$]; ~\vspace{.1in}
	
	$\triangleright$ {\color{gray} Compute the importance of image \& speech}\\
    $score\_1$ = (($pred\_1$ - $pred_b$) + ($pred_f$ - $pred_2$))/2\;\\
    $score\_2$ = (($pred\_2$ - $pred_b$) + ($pred_f$ - $pred_1$))/2~\vspace{.1in}
    
    $\triangleright$ {\color{gray} Placeholders for speech \& image shap values}\\
    $SHAP\_value\_img$ = np.zeros([$wd,\ ht$])\;\\
    $SHAP\_value\_speech$ = np.zeros([$wd,\ ht$])\;\\
    $times$ = $times$ - 1 ~\vspace{.1in}
    
    $\triangleright$ {\color{gray} Compute $SHAP\_value\_img$ and $SHAP\_value\_speech$}  \\
    Compute $SHAP\_value\_img$ using Eq. \ref{eq:eq2}, \ref{eq:eq3} and \ref{eq:eq5}\;\\
    Compute $SHAP\_value\_speech$ using Eq. \ref{eq:eq2}, \ref{eq:eq3} and \ref{eq:eq5}\;\\
	
	\caption{\noindent Proposed interpretability technique}
	\label{algo:a1} 
\end{algorithm} 
}

\section{Experiments}\label{sec:experiments}	
\subsection{Experimental setup}
The proposed system's network has been trained using Nvidia Quadro P5000 Graphics Card, whereas 64 bit Core(TM) i7-8700 Ubuntu system with 3.70 GHz 16GB RAM has been used for model evaluation.

\subsection{Training strategy}\label{sec:dataset}
The model has been trained using a batch size of 64, a train-test split of 70-30, 5-fold cross-validation, \textit{Adam} optimizer, \textit{ReLU} activation function with a learning rate of $8 \times 10^{-6}$. The baselines and proposed models converged regarding validation loss in 18-23 epochs. The models have been trained for 30 epochs as a safe upper bound. A weighted combination of categorical cross entropy with weights 1 and 0.5 and categorical focal loss~\cite{lin2017focal} has been used as the loss function. \textit{EarlyStopping} and \textit{ReduceLROnPlateau} have been incorporated with patience values 5 and 2. Accuracy, macro f1 \cite{opitz2019macro}, and \textit{CohenKappa} \cite{vieira2010cohen} have been analyzed for evaluation. 

\subsection{Ablation studies and models}\label{sec:ablation}
The following studies analyze the effect of using multimodal information and various network configurations.

\subsubsection{Effect of multiple modalities}
We first worked on SER and IER alone, using only speech samples and images from the IIT-R SIER dataset. Then we combined the information from speech \& image modalities and performed multimodal emotion recognition. The IER-only experiments demonstrated high training but low validation accuracy. The convergence of accuracy and f1 score was not in line, and \textit{CohenKappa} metric's value was low, denoting over-fitting for a particular class. The accuracy and f1 score converged in line for SER-only experiments, though the accuracy was less.

\subsubsection{Effect of various network configurations}
As depicted in Fig. \ref{fig:archi}, ParallelNet consists of a family of networks where $N1$ and $N2$ can be varied in different situations. We first keep $N2$ fixed as EfficientNet \cite{tan2019efficientnet} and evaluate three configurations for $N1$ -- Configuration 1 uses two criss-crosses before and after $N2$. A \textit{criss-cross} is a position combining two different modalities' networks. Configuration 2 \& 3 implement single criss-cross before and after $N2$. Three baseline models have been implemented in line with these configurations. Configuration 3 was chosen for final implementation as it shows in-line convergence \& improved performance.\vspace{.1in}

Further, keeping Configuration 3 fixed for $N1$'s configuration, following choices have been evaluated for $N2$ -- VGG~\cite{simonyan2014very} (VGG-16, VGG-19), ResNet~\cite{he2016deep} (ResNet-34, ResNet-50, ResNet-101), InceptionNet~\cite{szegedy2015going} (Inception 3a, Inception 4a), MobileNet~\cite{howard2017mobilenets} and DenseNet~\cite{huang2017densely}. The best performance has been observed with VGG16 as $N2_s$ and VGG19 as $N1_i$, which have finally been implemented by the `ParallelNet.' The baseline \& proposed models determined through the aforementioned studies are listed below, and their performance in terms of validation accuracies have been summarized in Table~\ref{tab:ablation}. \vspace{.1in}  

\begin{itemize}
	\item \textbf{Baseline 1} -- $N1$: Two criss-cross, $N2$: EfficientNet. It divides $N1$ into two parts and uses two criss-crosses before and after $N2$. A \textit{criss-cross} is a position combining two different modalities' networks. \vspace{.1in}
	\item \textbf{Baseline 2} -- $N1$: Criss-cross before $N2$; $N2$: EfficientNet. \vspace{.1in}
	\item \textbf{Baseline 3} -- $N1$: Criss-cross after $N2$; $N2$: EfficientNet.  \vspace{.1in}
	\item \textbf{Proposed} -- $N1$: Criss-cross after $N2$; $N2$: VGG.
\end{itemize}

\begin{table}[!h]
	\centering
	{
		\caption{Ablation studies' summary.}
		\label{tab:ablation}
		\resizebox{.5\textwidth}{!}
		{%
			\begin{tabular}{lc}
				\hline
				\textbf{Model} & \textbf{Accuracy} \\ \hline
				SER Only    & 60.17\% \\
				IER Only    & 66.93\% \\  \hdashline
				Baseline 1  & 63.93\% \\ 
				Baseline 2  & 61.81\% \\ 
				Baseline 3  & 67.70\% \\ \hdashline 
				Proposed (`ParallelNet')  & \textbf{83.29}\% \\ \bottomrule 
			\end{tabular}  
		}
	}
\end{table}

\section{Results and discussion}\label{sec:results} 
The emotion classification results have been discussed in this Section, along with their interpretation and a comparison of sentiment classification results with existing methods.

\subsection{Quantitative results}\label{sec:acc}
\begin{figure}[!h]
	\centering
	\includegraphics[width=0.53\textwidth]{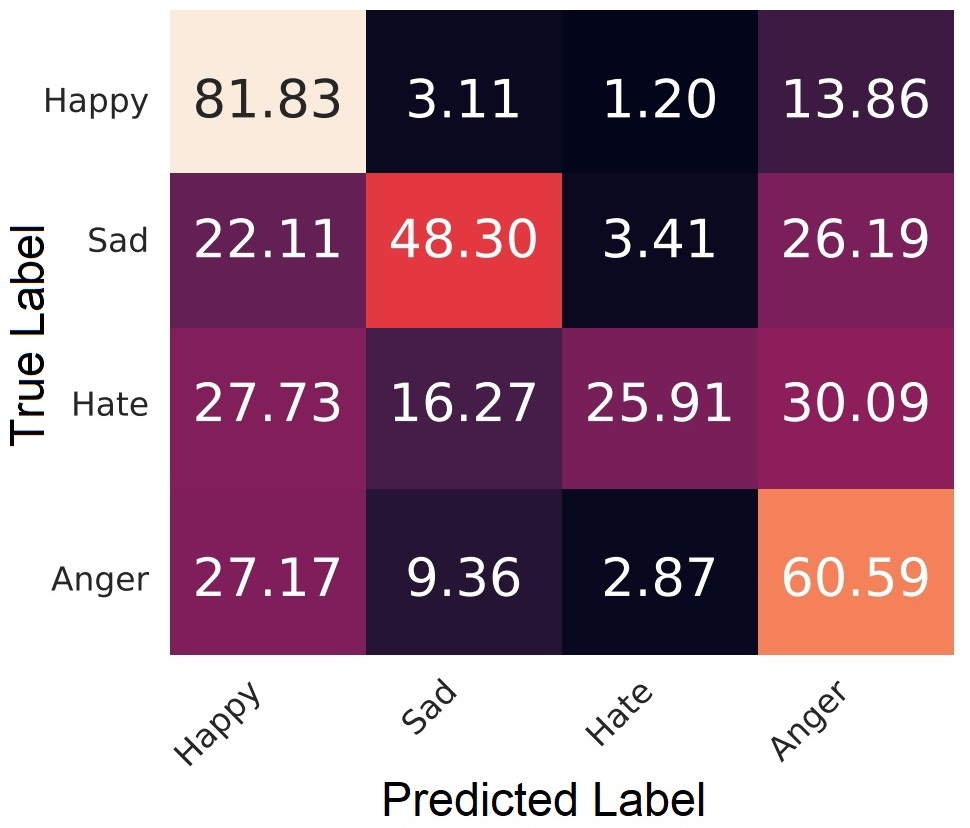}\vspace{.1in}
	\caption{Confusion matrix showing class-wise accuracies.} 
	\label{fig:confmat} \vspace{-.2in}
\end{figure}

The `ParallelNet' has achieved emotion recognition accuracy of \textbf{83.29}\%. Its class-wise accuracies are shown in Fig.~\ref{fig:confmat}.

\begin{figure}[!h]
	\centering
	\captionsetup{justification=centering}
 	\subfloat[Sample Result 1]
	{\includegraphics[width=.9\textwidth]{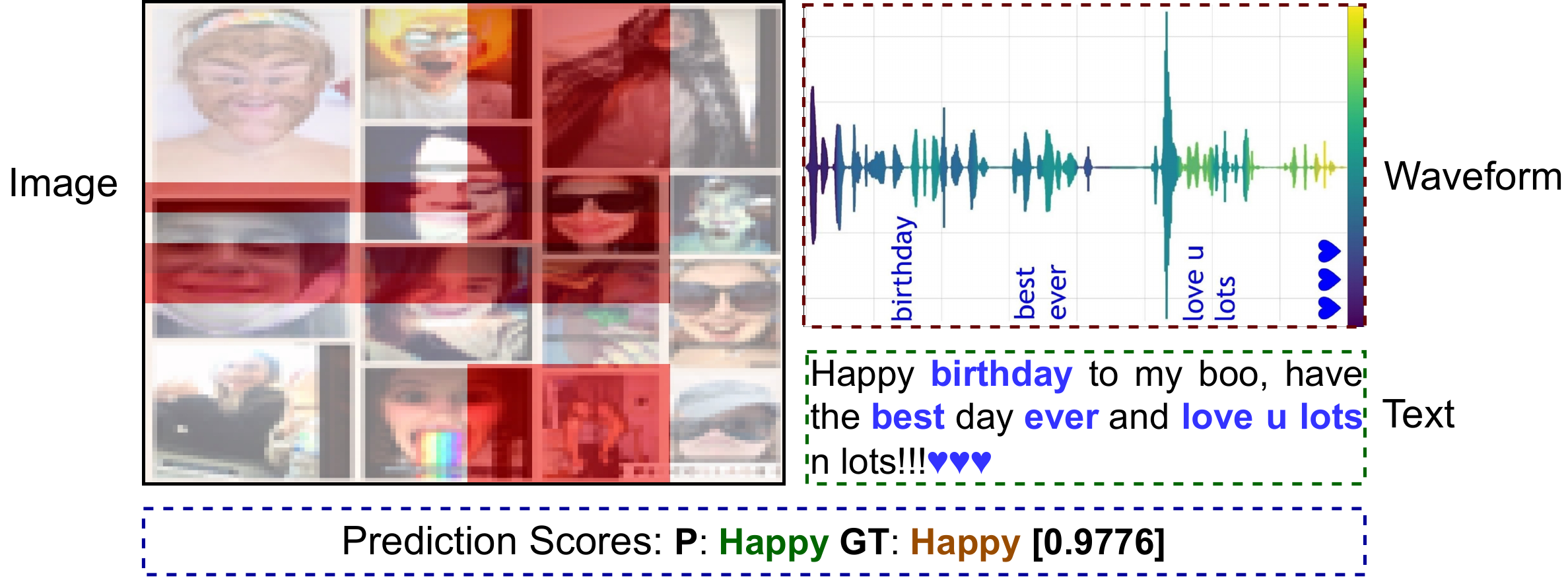}
		\label{fig:f1}
	}\vspace{-.1in}
 	\subfloat[Sample Result 2]
	{\includegraphics[width=.9\textwidth]{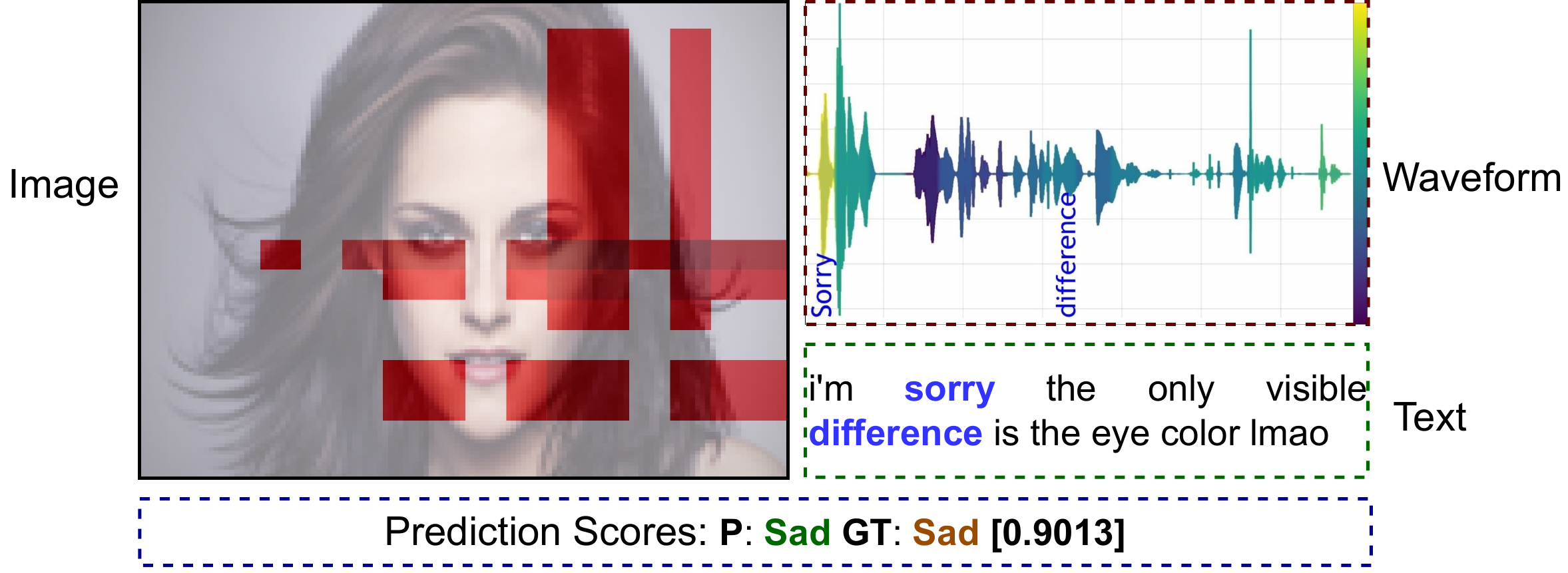}
		\label{fig:f2}
	}\vspace{-.1in} 
 	\subfloat[Sample Result 3]
	{\includegraphics[width=.9\textwidth]{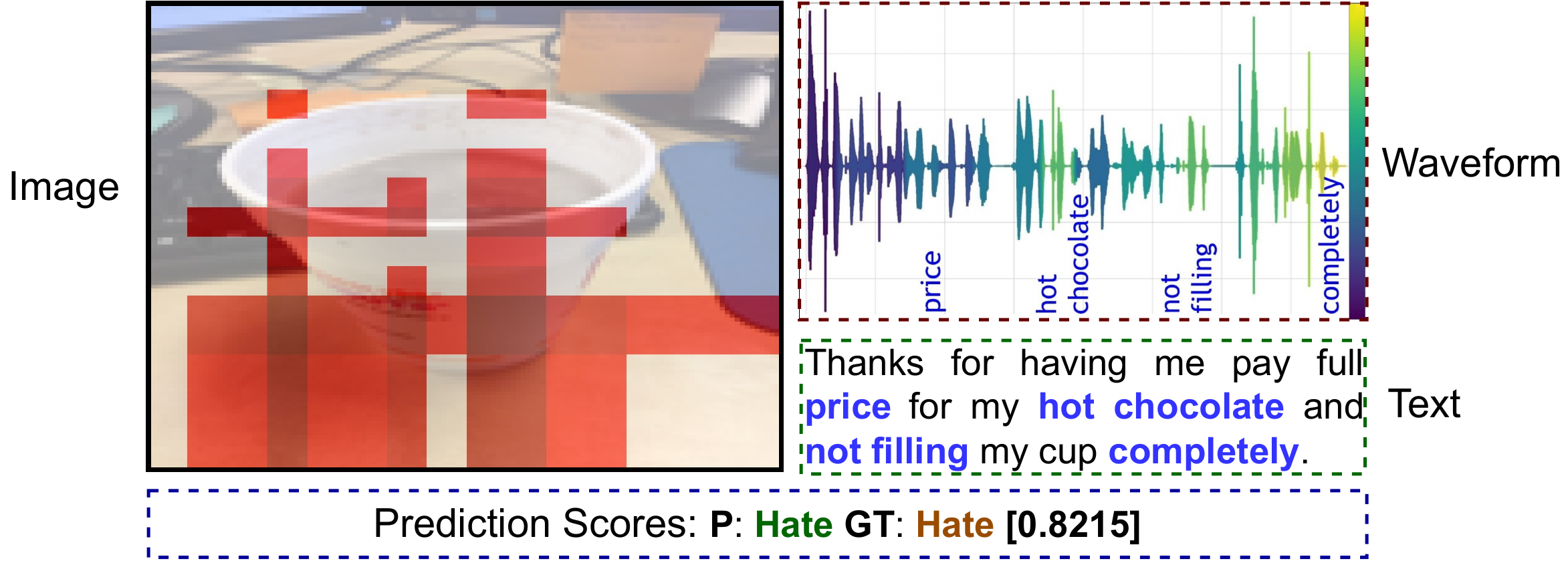}
		\label{fig:f3}
	}\vspace{-.1in} 
 	\subfloat[Sample Result 4]
	{\includegraphics[width=.9\textwidth]{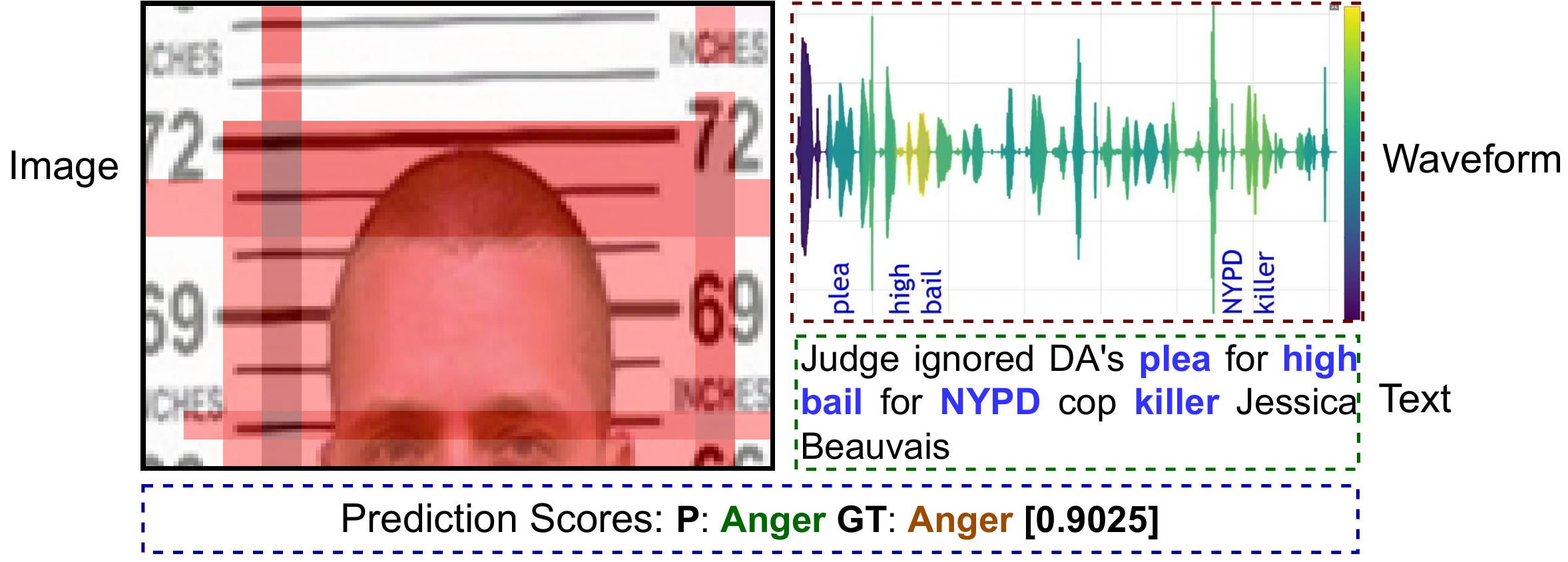}
		\label{fig:f4}
	} 
	\caption{Sample results; here, `P', `GT' and `Score' denote the predicted label, ground-truth label and softmax score.}
	\label{fig:sample_res}
\end{figure}

\subsection{Qualitative results}	 
Fig.~\ref{fig:sample_res} shows sample emotion classification \& interpretation results. The important speech and image features contributing to emotion classification are obtained, and corresponding words are highlighted. In the waveform, yellow and blue correspond to the most and least important features.  

\subsection{Results comparison}
\textbf{Comparison with existing Sentiment Analysis methods}: The emotion recognition results have been reported in Section~\ref{sec:acc}. The IIT-R SIER dataset has been constructed from the B-T4SA dataset in this paper; hence, there are no existing emotion recognition results for it. However, sentiment classification (into neutral, negative, and positive classes) results on the B-T4SA dataset are available in the literature, which has been compared with the proposed method's sentiment classification results in Table~\ref{tab:sota}.


\begin{table}[!h]
	\centering
	{
		\caption{Comparing existing sentiment analysis methods}
		\label{tab:sota}
		\resizebox{.82\textwidth}{!}
		{%
			\begin{tabular}{@{}lll@{}}
				\toprule
				\textbf{Approach} & \textbf{Author}   & \multicolumn{1}{c}{\textbf{Accuracy}} \\ \midrule
				Cross-Modal Learning & Vadicamo et al.~\cite{Vadicamo_2017_ICCVW} & 51.30\% \\
				Multimodal Sentiment Analysis & Gaspar et al.~\cite{gaspar2019multimodal} & 60.42\% \\
				Hybrid Fusion & Kumar et al.~\cite{kumar2021hybrid} & 86.70\% \\
				\hdashline
				\multicolumn{2}{c}{ParallelNet (Proposed)} & \textbf{89.68}\% \\ \bottomrule 
			\end{tabular}
	    }
	} 
\end{table}

\noindent \textbf{Comparison with human evaluation}: On considering the multimodal context from image and speech modalities, the human evaluation (See Table \ref{tab:human_eval}) and automatic evaluation (using ParallelNet. See Table \ref{tab:ablation}) resulted in emotion classification accuracies of 80.46\% and 89.68\% respectively. In both cases, the emotion classification performance improved on considering the multimodal context compared to considering only image or speech modality. It establishes the importance of considering complementary information from multiple modalities for emotion recognition.

\subsection{Discussion}
The proposed system classifies a given multimodal input containing speech \& the corresponding image into `anger,' `happy,' `hate,' and `sad' classes. The proposed interpretability technique identifies the important speech \& image features contributing to emotion recognition. 
An alternate procedure to construct the IITR-SIER dataset was to retain only those samples from the BT4SA dataset for which SER \& IER models predicted the same label and discard the rest of the samples. However, it would have caused a bias towards the models used in the first place for creating these labels. The SER \& IER models have been retrained on the IITR-SIER dataset instead of using the pre-trained weights of the models used to construct the IITR-SIER dataset. However, suppose somebody uses the pre-trained models of either one of the two modalities (trained on IEMOCAP and Flickr \& Instagram datasets, respectively) used during dataset construction. In that case, they will get a 100\% accuracy. The closest to them, any other evaluated machine learning model is, the more favorable its evaluation would be. That's why the proposed procedure of considering the prediction probabilities for all emotion classes is more effective in capturing the overall emotional context represented by both modalities in combination. It leads to generating more accurate ground-truth labels.\vspace{.1in} 
 
The ParallelNet's architecture has been determined through extensive ablation studies. It consists of a family of networks where $N1$ and $N2$ can be varied in different situations. We've first determined the optimal configuration for $N1$ to combine speech \& image modalities' information. Further, VGG, ResNet, InceptionNet, MobileNet, and DenseNet have been evaluated for $N2$. The best performance has been observed with VGG. The ResNet depicted very slow learning for a lower learning rate, while the learning fluctuated significantly for a higher learning rate. The model converged faster for the Inception Net and Efficient Net; however, the accuracy is lower. MobileNet and DenseNet have also resulted in low performance.\vspace{.1in}

Apart from the experimental validation in Table \ref{tab:ablation}, Fig.~\ref{fig:sample_res} qualitatively re-affirms the importance of combining complementary information from multiple modalities for more accurate emotion recognition. In the first \& second cases, the image and speech features (shown by yellow parts of the waveform and denoted by corresponding words in blue) contribute to predicting the emotion class `sad.' In the third \& fourth cases, the image features have not been precisely captured, and the images seem neutral. However, the corresponding speech features \textit{not filling} and \textit{killer} contribute towards hatred and anger intent, which leads to recognizing the `hate' and `anger' classes.
\section{Conclusions and future work}\label{sec:conclusion}
The importance of utilizing information from multiple modalities has been established for emotion recognition. The proposed system, \textit{ParallelNet}, has resulted in better performance than SER alone, IER alone, and baseline models. The proposed interpretability technique identifies the important image \& speech features contributing to emotion recognition.\vspace{.1in}

Future research plans include working on emotion recognition in other modalities such as text, videos, and emotion signal data. It is also planned to explore the interpretability of emotion recognition in the aforementioned modalities.

\section*{Declaration of Competing Interest} 
The authors declare that they have no known competing financial interests or personal relationships that could have appeared to influence the work reported in this paper. 

\bmhead{Acknowledgments}
Ministry of Education, India, has supported this work with grant no. 1-3146198040. It has been carried out at the Machine Intelligence Lab, IIT Roorkee, India.

\section*{Declarations}
\noindent \textbf{Funding}: This research was supported by Ministry of Human Resource Development (MHRD) INDIA with reference grant number: 1-3146198040.\vspace{.1in}

\noindent \textbf{Conflicts of interest}: Authors have no conflict of interest.\vspace{.1in}

\noindent \textbf{Code availability}: available at \href{https://github.com/MIntelligence-Group/SpeechImg_EmoRec}{https://github.com/MIntelligence-Group/SpeechImg_EmoRec}.\vspace{.1in}

\noindent \textbf{Availability of data and material}: available at  \href{https://github.com/MIntelligence-Group/SpeechImg_EmoRec}{https://github.com/MIntelligence-Group/SpeechImg_EmoRec}.\vspace{.1in}

\noindent \textbf{Authors' contributions}: \textit{Puneet Kumar}: Methodology, Implementation, Experiments, Result Analysis, Writing - original draft \& editing. \textit{Sarthak Malik}: Data Curation, Implementation, Conceptualization, Validation, Writing - review. \textit{Balasubramanian Raman}: Conceptualization, Writing - review, Supervision, Project administration.\vspace{.1in} 

\noindent \textbf{Ethics approval}: `Not applicable'.\vspace{.1in}

\noindent \textbf{Consent to participate}: `Not applicable'.\vspace{.1in}

\noindent \textbf{Consent for publication}: `Not applicable'.\vspace{.1in}

\noindent This article does not contain any studies with human participants or animals performed by any of the authors	

\bibliography{ref} 

\end{document}